*Research Article*

# Optimized Extreme Learning Machine for Power System Transient Stability Prediction Using Synchrophasors

**Yanjun Zhang,[1] Tie Li,[1] Guangyu Na,[1] Guoqing Li,[2] and Yang Li[2]**

[1]*State Grid Liaoning Electric Power Supply Co. Ltd., Shenyang 110006, China*
[2]*School of Electrical Engineering, Northeast Dianli University, Jilin 132012, China*

Correspondence should be addressed to Yang Li; 876464104@qq.com

A new optimized extreme learning machine- (ELM-) based method for power system transient stability prediction (TSP) using synchrophasors is presented in this paper. First, the input features symbolizing the transient stability of power systems are extracted from synchronized measurements. Then, an ELM classifier is employed to build the TSP model. And finally, the optimal parameters of the model are optimized by using the improved particle swarm optimization (IPSO) algorithm. The novelty of the proposal is in the fact that it improves the prediction performance of the ELM-based TSP model by using IPSO to optimize the parameters of the model with synchrophasors. And finally, based on the test results on both IEEE 39-bus system and a large-scale real power system, the correctness and validity of the presented approach are verified.

## 1. Introduction

Monitoring the power system stability status in real-time has been regarded as an important work to guarantee the power system safe and stable operation [1, 2]. Up to now, the existing transient stability analysis (TSA) methods mainly can be divided into 3 classes: direct methods [3], time-domain simulations [4], and the extended equal area criterion method [5]. Unfortunately, these methods cannot work well for real-time stability analysis of modern complex power systems.

In recent years, pattern-recognition-based TSA (PRTSA) has been attracting the ever-growing attention of researchers all over the world [6, 7]. This kind of method has proved to be potential in the area of on-line dynamic security analysis by applying of the techniques of machine learning. By far, the PRTSA model mainly includes artificial neural networks (ANN), decision trees (DT), and support vector machines (SVM) [8–15]. However, the reported PRTSA approaches usually suffer from some inherent disadvantages and lack the ability of big data management and utilization, which restricts its further application in actual operating scenarios. For example, ANN has problems of overfitting, local optima, and slow convergence, and SVM has difficulty in parameter selection. On the other hand, wide area measurement systems (WAMS) provide the synchronous measurement information for the wide area power systems [16], which makes it possible to explore wide area protection and control schemes to avoid the system collapse [17–19].

In recent years, a novel machine learning algorithm called extreme learning machine (ELM) is proposed by Huang et al. [20]. Contrasted with those conventional PRTSA approaches, ELM has a lot of significant advantages, such as better generalization ability and a much faster learning speed [21–23]. Inspired by the social behavior of flocks, particle swarm optimization (PSO) algorithm is proposed in 1995 [24]. PSO has been widely used to solve a variety of optimization problems with many of advantages including good robustness, fast convergence speed, and high search efficiency [25].

In this paper, a novel ELM-based transient stability prediction (TSP) method using synchronized measurements is proposed. Moreover, to further improve the prediction performance, the ideal model is obtained by applying the improved particle swarm optimization (IPSO) algorithm to select the optimal parameters of the model.

The rest of this paper is arranged as follows. First of all, the used methodologies including ELM classification, PSO are presented briefly. Secondly, the proposed real-time TSP method based on IPSO-ELM is presented in detail. Finally, the proposal is tested using the IEEE 39-bus system and a real system.

## 2. Related Methodologies

### 2.1. ELM Classification.
Assuming an ELM with $L$ hidden layer neurons to model data samples $\{\mathbf{x}_i, \mathbf{y}_i\}_{i=1}^{N}$, it can be mathematically represented as

$$\sum_{i=1}^{L} \boldsymbol{\beta}_i G(\mathbf{a}_i, b_i, \mathbf{x}_j) = \mathbf{y}_j, \quad j = 1, \ldots, N, \quad (1)$$

where $\mathbf{a}_i$ and $\boldsymbol{\beta}_i$ are, respectively, the input and output weights vector, $G(\cdot)$ is activation functions, and $b_i$ denotes the bias of the $i$th hidden node.

For the convenience of expression, (1) can be rewritten as

$$\mathbf{H}\boldsymbol{\beta} = \mathbf{Y},$$

$$\mathbf{H} = \begin{bmatrix} h_1(\mathbf{x}_1) & \cdots & h_L(\mathbf{x}_1) \\ \vdots & \vdots & \vdots \\ h_1(\mathbf{x}_N) & \cdots & h_L(\mathbf{x}_N) \end{bmatrix}, \quad (2)$$

where $\mathbf{H}$ is the hidden layer output matrix.

ELM is to minimize the training error as well as the norm of the output weights [20]

$$\text{Minimize} : \|\mathbf{H}\boldsymbol{\beta} - \mathbf{Y}\|_2 \text{ and } \|\boldsymbol{\beta}\|. \quad (3)$$

Finally, the minimal norm least square method is used in the original implementation of ELM

$$\widehat{\boldsymbol{\beta}} = \mathbf{H}^{\dagger}\mathbf{Y}, \quad (4)$$

where $\mathbf{H}^{\dagger}$ is the Moore-Penrose generalized inverse of $\mathbf{H}$.

### 2.2. PSO.
The fundamental principle of PSO is to find the optimal solution in the complex search-space by moving candidate solutions (called particles) according to the competition and collaboration among particles through repetitive iterations. The movement of each particle is determined by a mathematical formulae over its position and velocity. In the $k + 1$ iteration, the velocity and position renewal equation of $i$th particle are, respectively, as follows:

$$\begin{aligned} v_i^{k+1} &= wv_i^k + c_1 r_1 \left(P_{\text{best}} - s_i^k\right) + c_2 r_2 \left(G_{\text{best}} - s_i^k\right), \\ s_i^{k+1} &= s_i^k + v_i^{k+1}, \end{aligned} \quad (5)$$

where $P_{\text{best}}$ and $G_{\text{best}}$ are, respectively, denoted as the local and global best known solution; $k$ and $w$ are, respectively, denoted as the evolutionary generation and the inertia weight; $c_1$ and $c_2$ are the learning factors, which represent the self-cognition and social-cognition in turn; $r_1$ and $r_2$ are uniform random numbers obeying the 0-1 distribution.

## 3. Real-Time TSP Based on IPSO-ELM

### 3.1. IPSO Algorithm.
As pointed out by the famous "No free lunch" theorem, the overall performances of different optimization algorithms are equivalent [26], which implies that none of algorithms can always achieve the optimal for all aspects. In this paper, a mutation strategy is introduced to avoid the premature convergence to local optimum of PSO.

First, the optimization process is monitored by dynamically monitoring changes in population fitness variance $\sigma^2$:

$$\sigma^2 = \sum_{i=1}^{N_p} \left(\frac{f_i - f_{\text{avg}}}{f_{\text{best}}}\right)^2, \quad (6)$$

where $N_p$ denotes the population size, $f_i$ is the fitness value of the $i$th individual particle, $f_{\text{best}}$ refers to the best fitness value in the whole population, and $f_{\text{avg}}$ denotes the average fitness in the current iteration.

Second, when premature convergence occurs, a mutation strategy is used to maintain population diversity. Specifically, the positions of particles are updated by adding random perturbations timely, as shown as follows:

$$x^{(k+1)} = x^{(k)} + c_m \left(x_{\max} - x_{\min}\right) \cdot (\text{rand} - 0.5), \quad (7)$$

where $c_m$ is the variation coefficient, rand is a real number randomly generated in the range from 0 to 1, $x^{(k+1)}$ and $x^{(k)}$ are, respectively, the positions of the $(k+1)$th and $k$th iteration of particles.

The criterion to determine the occurrence of premature convergence is given as follows:

$$m < \frac{\sigma_{k+1}}{\sigma_k} < n. \quad (8)$$

Here, $\sigma_{k+1}$ and $\sigma_k$ are the population fitness variances of the $(k + 1)$th and $k$th iteration.

### 3.2. Fitness Function.
As is known, a proper fitness function plays an important role in optimization problems. In this paper, the fitness function is the classification accuracy of 5-fold cross-validation (CV):

$$F(\boldsymbol{\theta}) = A_{5\text{-CV}}, \quad (9)$$

where $\boldsymbol{\theta}$ is the model parameter vector to be optimized, which is represented by the position of each particle.

### 3.3. Coding Scheme.
A mixed-integer encoding scheme is used in the optimization process [27]. Here, it is considered that the $k$th individual particle/state $\boldsymbol{\theta}_k$ will be constituted by

$$\boldsymbol{\theta}_k = \left[\mathbf{a}_1^T, \ldots, \mathbf{a}_L^T, b_1, \ldots, b_L, s_1, \ldots, s_n, cf_1, \ldots, cf_h\right]^T, \quad (10)$$

where $n$ is the total count of input features, $s_i$ ($i = 1, \ldots, n$) is a binary variable, and each binary code ("1" or "0") refers to whether the corresponding feature is selected or not; $cf_j = \{0, 1, 2\}$ ($j = 1, \ldots, L$) is an integer variable that defines

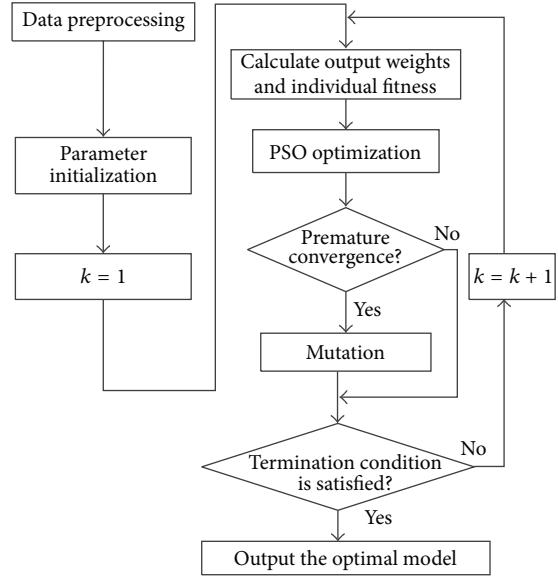

Figure 1: Flowchart of the modeling process.

the activation function $f_j$ of each neuron $j$ of the hidden layer as follows:

$$f_j(v) = \begin{cases} 0, & \text{if } cf_j = 0, \\ \dfrac{1}{(1 + \exp(-v))}, & \text{if } cf_j = 1, \\ v, & \text{if } cf_j = 2. \end{cases} \quad (11)$$

The use of parameters $cf_j$ makes possible the adjustment of the number of neurons (if $cf_j = 0$ the neuron is not considered) and the activation function of each neuron (sigmoid or linear function). To facilitate the optimization, the decision variables are mapped into real variables within the interval $[0, 1]$; and all variables need to be converted into their true value when computing the fitness value of each particle [27].

*3.4. Modeling Process.* The modeling process of the proposed method can be divided into 8 steps.

*Step 1.* The used data preprocessing approach here is $z$-score standardization method [15]:

$$z' = \frac{(z - \overline{Z})}{\sigma_Z}, \quad (12)$$

where $\overline{Z}$ is the mean of any feature $Z$ in sample data, $\sigma_Z$ is the standard deviation of the feather $Z$; $z'$ is the normalized value corresponding to $z$, $z \in Z$.

*Step 2.* Initialization of the parameters: the maximum iteration number is assigned to 200, the population size $N_p$ is set to 20, and the number of ELM hidden layer neurons is 50.

*Step 3.* Initialization of the population: $N_p$ solutions are generated randomly, and each solution is corresponding to a particle, which is encoded according to (10).

*Step 4.* According to (4), calculate the output weights $\beta$ and the individual fitness values in turn.

*Step 5.* According to the optimization mechanism of PSO algorithm, update the location of individual particle and generate the next populations.

*Step 6.* Dynamic monitoring of changes in the population fitness variance: once premature convergence occurs, save the current optimal solution $\theta^*$, and go on to the mutation operation. If a better solution $\theta'$ ($F(\theta') > F(\theta^*)$) is found in the solution space, then update the optimal solution $\theta^* = \theta'$, and quit the mutation operation.

*Step 7.* Judgment of termination condition: the optimization process will be terminated, if the current number of iterations $k$ exceeds the prespecified maximum number of iterations or the value of fitness function is greater than 99.00%; otherwise, $k = k + 1$ and jump to Step 4.

*Step 8.* Acquisition of the ideal model: output the optimal solution $\theta^*$, and obtain the ideal TSP model.

The flowchart of the modeling process is shown in Figure 1.

*3.5. Construction of the Initial Feature Set.* As is known to us all, input features play an important role in PRTSA [8, 14, 15]. However, the used features in previous works are mainly prefault static features. The reason for this is that the former measurement systems are not able to provide wide area dynamic information. To take full advantage of the synchrophasors information from WAMS [16–19], the presented approach selects input features from both prefault static information and postfault dynamic information.

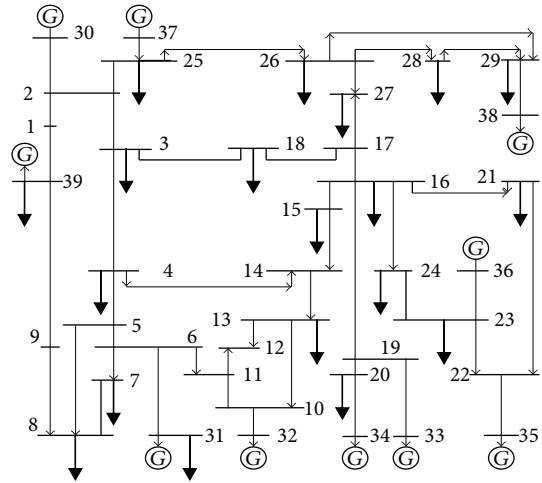

Figure 2: IEEE 39-bus test system.

As an extension of related works, the selected initial features used in the presented approach are the same as the ones in [15] (see [15] for further details). These features are be made up of prefault static features and the postfault dynamic features. Therefore, the above selected feathers comprehensively indicate the stability of power systems during different stages of the disturbance process, and they are appropriately selected to constitute the original feature set of the transient disturbed pattern space.

*3.6. TSP Based on the Trained Model.* In this section, how the ELM-based TSP model is used after it has been trained is explained in brief. It should be noted that synchrophasors are simulated through the detailed time-domain simulations in this work. Here, all the used input features can be obtained from the following physical quantities, comprising the rotor angle, angular velocity, mechanical power and electromagnetic power of each generator, and the generators' inertial time constants; and all these physical quantities are available from PMU measurements except for the given inertial time constants. Therefore, the proposed method is able to be applied to TSP based on PMU measurements.

In the present approach, it is supposed that tripping signal(s) issued by the local protection is available for triggering the TSP system. Once a fault is cleared by the action of relevant relays, the trigger allows starting of taking the samples of the input variables to construct the input vector for the proposed ELM-based TSP model. And then, the proposed model takes 9 consecutive synchronously measured samples of each generator at the rate of 60 per sec to form the input vector for the classifier. Finally, for a specific input vector, the transient stability status of the disturbed power system can be immediately predicted by using the trained model.

## 4. Results and Discussion

For power system TSA, the New England 39-bus test system is a widely used test system to examine the performance of various assessment methods [9–15]. The system contains 10 generators and 39 buses, which is shown in Figure 2.

*4.1. Generation of Knowledge Base.* It is known that for PRTSA, the generalization ability of a TSA model largely depends on the completeness and representativeness of the knowledge base (KB). Hence, it is an important work to generate the used KB, reflecting the relationship between the input features and the stability status. In this work, KB is made through extensive time-domain simulations in detail. The employed generator model is four-order model with IEEE DC1 excitation system; the load model is the constant impedance model. The fault type is three-phase short-circuit faults, and the fault clearing times are varied from five to ten cycles. It is assumed that reclosures are successfully applied and the network topology is not changed when the fault is cleared. The load levels used ranged from 80% to 130% of the basic load level, and the active and reactive power outputs of each generator are correspondingly assigned. Among the total 3300 created samples, 2200 ones are randomly chosen as the training samples and the rest as the testing samples.

A class label Class_Lable of each sample is denoted by a transient stability index which is related to the relative rotor angle deviation during the transient period of a disturbed power system [13, 15]. The label Class_Lable of a sample is determined as

$$\text{Class\_Lable} = \text{sgn}\left(360° - |\Delta\delta|_{\max}\right), \quad (13)$$

where sgn(·) is a sign function, | · | is the absolute value function, and $\Delta\delta_{\max}$ is of the maximum relative rotor angle deviation between generators in the period. By plotting the rotor angle swing curves of the generators, an unstable case is illustrated in Figure 3, and a transient stable case is illustrated in Figure 4.

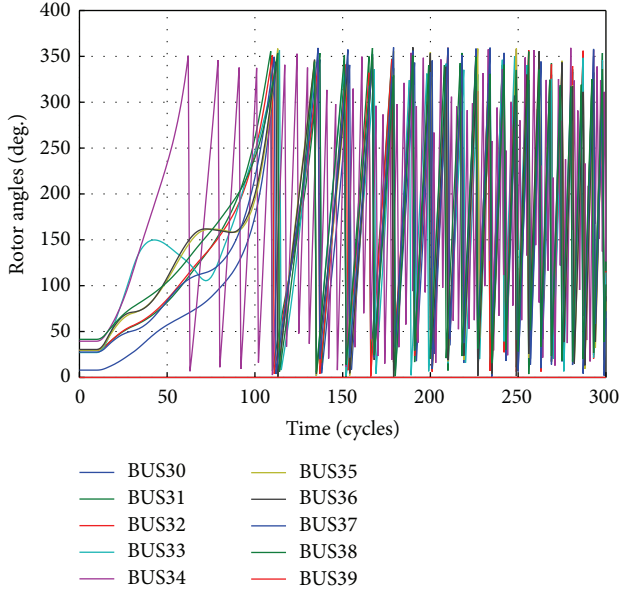

Figure 3: A transient unstable case.

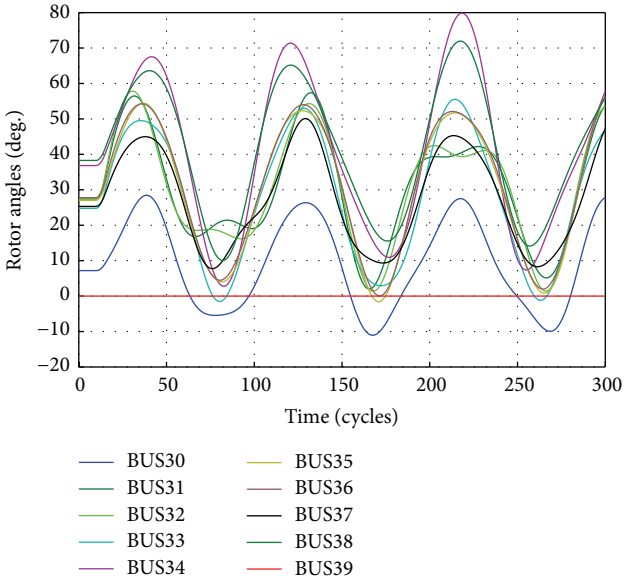

Figure 4: A transient stable case.

Table 1: Comparison results of different algorithms.

| Algorithm | Training time/s | Optimal fitness value/% | Success rate/% | Number of hidden nodes |
|---|---|---|---|---|
| IPSO | 16.52 | 97.50 | 92 | 18.35 |
| GA | 54.38 | 95.24 | 73 | 24.05 |
| PSO | 25.47 | 94.20 | 61 | 29.00 |

### 4.2. Training Results of Different Optimization Algorithms.

Comparison tests are carried out by using other optimization algorithms comprising PSO and genetic algorithm (GA). In order to facilitate comparative analysis, all the common parameters of these algorithms are assigned to the same numerical values; and other parameters are set as follows: in GA, the mutation probability is set to 0.01, and the crossover probability is assigned to 0.85; in PSO, both the two learning factors $c_1$ and $c_2$ are assigned to 2 and the inertia weight $\omega$ is set to be linearly decreased from 0.9 to 0.4. At the same time, taking into account the randomness of the above-mentioned algorithms, all of them are executed 100 iterations, independently. The comparison results of different optimization algorithms are listed in Table 1.

From Table 1, it can be seen that the proposed IPSO algorithm has better quality of solution, shorter training time, and higher search success rate than other optimization algorithms such as GA and PSO. The reason for this is that the dynamic monitoring mechanism and the mutation strategy are comprehensively employed during the optimization process in the proposed approach; thus it has the best result and the most stable performance. Therefore, it can be drawn that IPSO is able to solve the ELM's parameter optimization problem effectively. It should be noted that both the training time and the number of hidden nodes are the average value of 100 times in Table 1.

The fitness evolution curves of the optimal individuals for the three optimization algorithms during the process are shown in Figure 5.

Figure 5 shows that all these algorithms have obvious effects in the parameter optimization process for ELM. Among all these algorithms, IPSO has the fast search efficiency, which reaches the optimal solution only through 106 iterations; moreover, the best fitness value of IPSO is the highest one among all these optimization algorithms. At the same time, IPSO has a transient pause at 55th iteration but soon continues to decline. This suggests that IPSO can quickly jump out the local optimal solution and overcome premature phenomenon effectively for its powerful global search capability. Therefore, the above results demonstrate that IPSO is able to enhance the solution quality and search efficiency for the proposed TSP approach.

### 4.3. Test Results.
First, the proposed IPSO-ELM-based TSP model is tested. It is known that fast prediction of instability is crucial for TSA, since the transient stability is a very fast process which demands a control measure within short period of time (<1 s) [13]. The required observation time in the presented approach is 150 ms (9 cycles), and that allows over 400 ms for measurement, telecommunication, and processing delays. Once the data are in the control center, the prediction time using the trained model is short. For example, with a MATLAB code implemented on a PC with 2.66 GHz CPU and 3 GB of RAM, this calculation required only 42.267 ms. In practical applications, the computing time can be further reduced by means of using the state-of-the-art parallel computing and distributed computing technology. Therefore, the conclusion can be drawn on the basis of the evidence that the proposed method is able to predict the transient stability status of power systems in real-time.

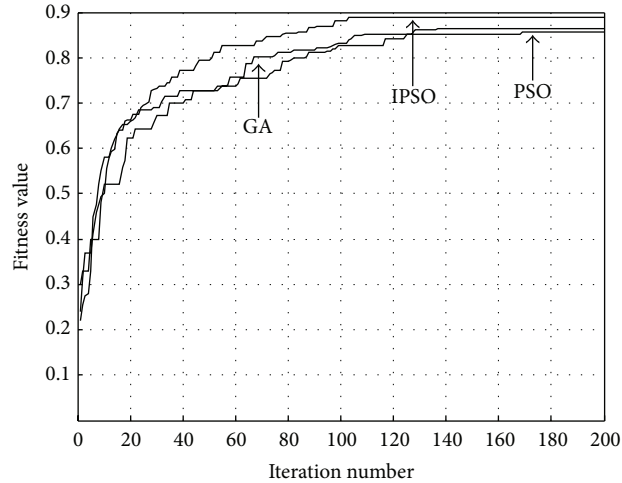

Figure 5: Best fitness evolution curves of different algorithms.

Table 2: Test results of the proposed approach.

| TSP model | Acc/% | Kap | AUC | $\eta$ |
|---|---|---|---|---|
| IPSO-ELM | 97.09 | 0.969 | 0.979 | 0.973 |
| ELM | 95.73 | 0.919 | 0.953 | 0.943 |

Table 3: Test results of other TSP models.

| TSP model | Acc/% | Kap | AUC | $\eta$ | Training time/s |
|---|---|---|---|---|---|
| DT | 94.82 | 0.908 | 0.950 | 0.935 | 28.34 |
| SVM | 95.64 | 0.925 | 0.964 | 0.949 | 47.15 |
| MLP | 94.09 | 0.896 | 0.946 | 0.928 | 97.58 |

And then, comparison tests by using ELM are carried out with the results shown in Table 2. Taking into account the occasionality of test accuracy Acc, the classification performance of the tested TSP model should also be evaluated by using some statistical indicators, such as the Kappa statistic value Kap and the area under the ROC curve AUC [15]. For this reason, a composite indicator $\eta$ is adopted here to comprehensively evaluate TSP models, as defined by the following equation:

$$\eta = \frac{\text{Acc} + \text{Kap} + \text{AUC}}{3}. \qquad (14)$$

Table 2 illustrates that the proposed algorithm manages to predict the transient stability with good accuracy. The reasons for this are as follows: on one hand, the classification features extracted from WAMS information fully reflect the dynamic response characteristics of the actual system; on the other hand, ELM has good generalization ability, for it pursues the minimization in both the norm of the output weights and the training error. At the same time, it can be observed that the classification ability of IPSO-ELM is much better than that of ELM. Therefore, the conclusions can be safely drawn that IPSO is an effective way to enhance the prediction performance for the proposed ELM-based TSP model.

*4.4. Results of Comparative Tests.* In order to further properly test the proposed scheme, comparative tests are carried out by using other TSP models, such as DT [7, 10], multilayer perception (MLP) [6, 9], and SVM [13], which are carried out with the results shown in Table 3.

The parameters of the above-mentioned TSP models are set as follows. DT is constructed using the C4.5 algorithm; the radial basis function is used as the kernel function of SVM, and its associated parameters are optimized through the grid search combined with 5-fold cross validation [13, 15]; in MLP, the hidden neuron number is set to 25, the learning algorithm is the backpropagation algorithm.

Table 3 shows that the proposed method has the superior predictive performance and training speed than other TSP models, not only because of the superior generalization ability and fast training speed of ELM itself but also because of application of IPSO to determine the optimal parameters of ELM. As a result, a conclusion can be safely drawn that the proposed method is effective in real-time transient stability prediction for power systems.

## 5. Application to the Power System of Liaoning Province

In order to further verify the applicability of the proposed approach to practical large power systems, the proposed method is examined on the power system of Liaoning province, China.

The modeled system comprises 91 generators, 750 major buses, and some series compensated lines and SVCs. It is a highly interconnected grid with an approximate installed capacity of 39657.2 MW, covering an area of 148,000 square kilometers. The system has formed 5 connected channels with the external power network.

Table 4: Test results of the power system of Liaoning province.

| TSP model | Acc/% | Kap | AUC | $\eta$ |
| --- | --- | --- | --- | --- |
| IPSO-ELM | 96.47 | 0.925 | 0.962 | 0.951 |

The contingencies considered are three-phase short-circuit faults. The stability criterion employed here is exactly the same as in the former case, IEEE 39-bus system. Through large amounts of simulations, there are 2000 samples created totally. Of all the samples, 1320 ones are selected as the training samples randomly, and the rest are used as the testing ones. The results of the power system of Liaoning province are shown in Table 4.

As demonstrated in Table 4, the proposed approach is applicable to large-scale real power systems as well. Furthermore, the results also show that the proposed approach is able to determine the transient stability status in the power system of Liaoning province, China. Therefore, the applicability of the proposal to a real power system is verified.

## 6. Conclusions

Machine learning has proved to be promising for solving on-line TSA problems. However, the existing PRTSA approaches cannot meet the needs of big data management and utilization. To overcome this problem, a new optimized ELM-based approach for real-time power system TSP using synchrophasors is presented. Based on the test results on the well-studied IEEE 39-bus system and a large-scale real power system, the conclusions can be obtained as follows.

(1) The method proposed is able to effectively predict the power system transient stability status using synchronized measurements, and it has better predictive performance and generalization ability than other commonly used TSP models, such as DT, MLP, and SVM.

(2) The predictive performance of the proposal is evidently strengthened by using the proposed IPSO to optimize the parameters of the ELM-based TSP model with synchronized measurements. Furthermore, by means of the introduction of the dynamic monitoring mechanism and the mutation strategy, IPSO has the better global optimization ability and the faster searching efficiency than the traditional optimization algorithms comprising GA and PSO.

(3) For a wide area protection and control system, the presented method may be used as trigger to start the related emergency control measures. Meanwhile, the proposed TSP model is able to be applied to other similar pattern recognition problems.

## Conflict of Interests

The authors declare that there is no conflict of interests regarding the publication of this paper.

## Acknowledgments

This research is supported by the Science and Technology Project of State Grid Corporation of China under Grant no. 2014GW-05, the key technology research of a flexible ring network controller and its demonstration application.

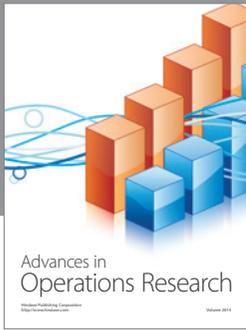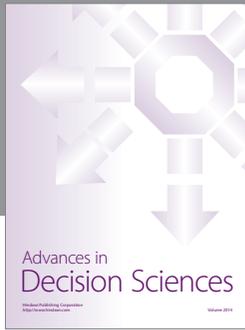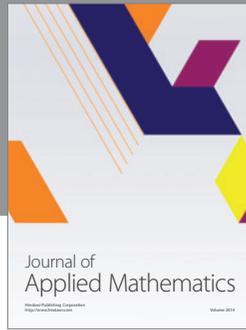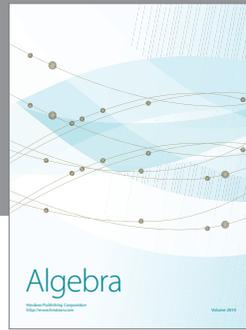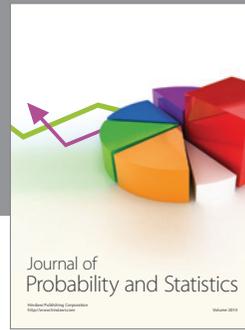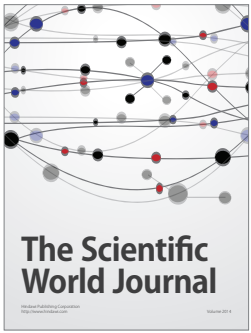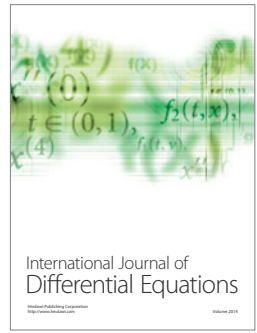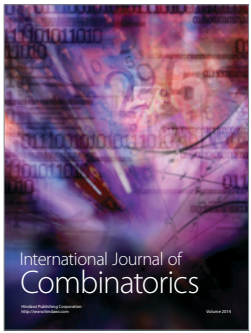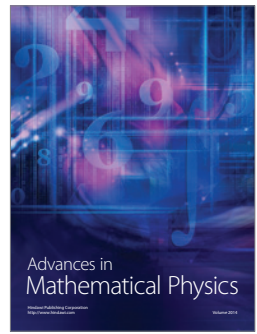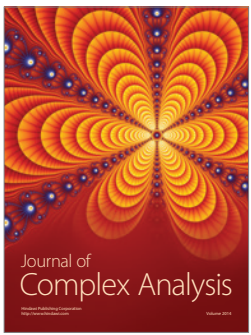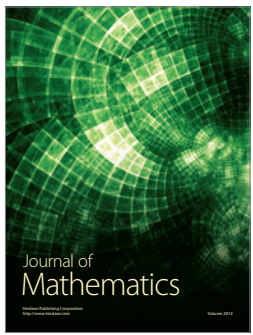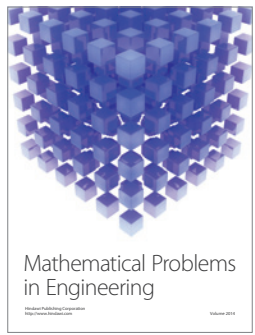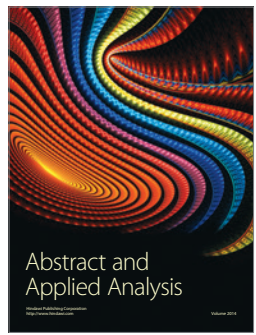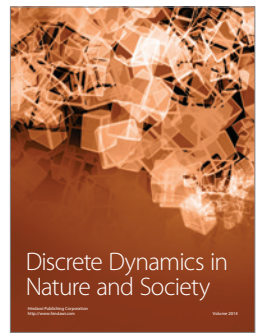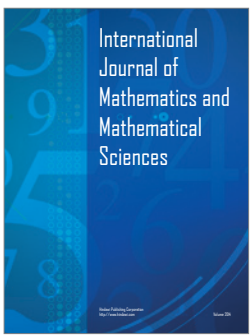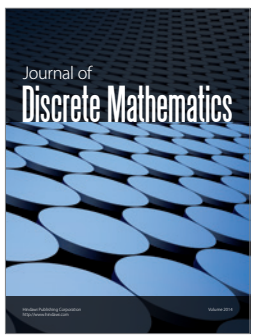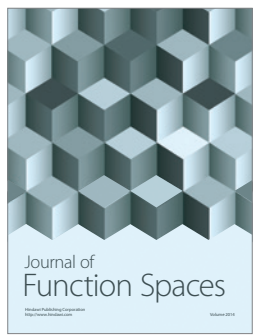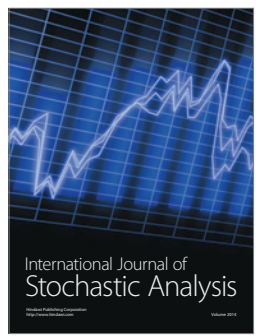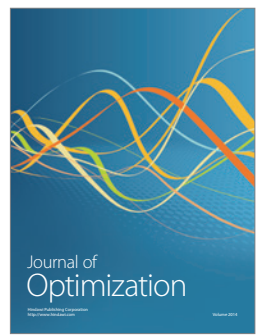